# New Worst-Case Upper Bound for #XSAT

Junping Zhou, Minghao Yin

**Department of Computer, Northeast Normal University, Changchun, China, 130117**

ymh@nenu.edu.cn

## Abstract

An algorithm running in $O(1.1995^n)$ is presented for counting models for exact satisfiability formulae(#XSAT). This is faster than the previously best algorithm which runs in $O(1.2190^n)$. In order to improve the efficiency of the algorithm, a new principle, i.e. the common literals principle, is addressed to simplify formulae. This allows us to eliminate more common literals. In addition, we firstly inject the resolution principles into solving #XSAT problem, and therefore this further improves the efficiency of the algorithm.

## Introduction

Tremendous efforts have been made on analyzing algorithms for difficult problems, such as propositional satisfiability (SAT) or model counting (#SAT). If P ≠ NP, these problems are all super-polynomial. When constructing the super-polynomial algorithms, improvements in the exponential time bounds are crucial in determining the size of these problems that can be solved, for even a slight improvement from $O(c^k)$ to $O((c-\varepsilon)^k)$ may significantly increase the size of these problems being tractable. Take the 3-SAT problem for example. The currently fastest deterministic algorithm for solving 3-SAT (Kutzkov et al. 2010) ran in $O(1.439^n)$, which is a meaningful advance over $O(2^n)$. And the 3-SAT instances with 65 variables can be solved by the algorithm in approximately $10^{10}$ steps, instead of $10^{19}$ (which may not be tractable). Therefore, it is significant to improve the upper bounds on the worst-case running time for problems with high computational complexity.

#XSAT is one of hard problems whose computational complexity is further up the polynomial hierarchy. Valiant (1979) has proved that #XSAT is #P-complete. This is a problem of computing the number of models for a given formula in Conjunction Normal Form (CNF), i.e., the number of distinct complete truth assignments to variables such that exactly one literal in each clause evaluates to *true*. In fact, the problem is a vital variant of the well-known #SAT problem, which has a wide range of applications, such as various probabilistic inference problems can be translated into #SAT problem (cf. Park 2002; Sang et al. 2005).

Recently, most of the efforts in algorithm construction have been dedicated to algorithms for #XSAT problem. For example, based on an algorithm for counting all maximum weight independent sets in a simple graph, Dahllof and Jonsson (2002) presented an upper time bound for #XSAT ($O(1.7548^n)$, where $n$ is the number of the variables). Further improved algorithm in (Porschen 2005) proposed a new upper time bound for the #XSAT ($O(1.3248^n)$). Instead of only outputting the number of models, Porschen in 2006 presented an algorithm solving #XSAT for enumerating all exact models of the input formula. By addressing a simple reduction, Dahllof et al. (2004) presented a #XSAT algorithm which ran in $O(1.2190^n)$, which is the best upper bound so far.

The aim of this paper is to exploit new upper bounds for #XSAT. We provide an algorithm for solving #XSAT. This algorithm employs a new principle, i.e. the common variables principle, to simplify formulae. This allows us to eliminate more common variables, and therefore improves the efficiency of the algorithm. In addition, we firstly inject the resolution principles into solving #XSAT problem, which further improves the efficiency of the algorithm. By analyzing the algorithm, we present a deep analysis and obtain the worst-case upper bound $O(1.1995^n)$ for #XSAT.

## Problem Definitions

We describe some definitions used in this paper. Let $V$ be a set of propositional variables. Each variable can take the values *true* or *false*. A literal is either a variable $x$ or its negation $\neg x$. Also, each literal can take the values *true* or *false*. If a literal is $l$, the negation of the literal is $\neg l$. The logic disjunction of a finite number of literals forms a clause. A clause containing exactly $k$ literals is also called *k-clause*. The length of a clause is the number of literals in it, denoted by $|C|$. A clause $C$ is a unit clause if the length of the clause is 1. And if there is a clause $x_1 \vee x_2 \vee ... \vee x_i \vee C$ such that $C$ is a disjunction of one or more literals, we call $C$ is a sub-clause. A formula $F$ in Conjunction Normal Form (CNF) is a logic conjunction of a set of clauses. A variable occurring once in $F$ is referred to as singleton. The degree of a variable $x$, represented by $\varphi(x)$, is the number of times it occurs in $F$. The degree of a formula $F$, denoted by $\varphi(F)$, is the maximum degree of variables in $F$. A literal $l$ is an $(i^+, j^+)$-literal if $F$ contains at least $i$ occurrences of $l$ and at least $j$ occurrences of $\neg l$. A literal $l$ is an $(i, j)$-literal if $F$ contains exactly $i$ occurrences of $l$ and exactly $j$ occurrences of $\neg l$. And a literal $l$ is monotone if its complementary literal does not appear in $F$. A truth assignment for $F$ is a map that assigns each variable a value. When a truth assignment makes the $F$ *true*, we say that the truth assignment satisfies $F$. The *exact satisfiability problem* (XSAT) is to determine if a CNF-formula has a truth assignment satisfying exactly one literal in each clause. The satisfying assignment of the XSAT, called as a model, is a truth assignment such exactly one literal in each clause is *true*. The counting exact satisfiablity problem

(#XSAT) is to compute the number of models for a formula. We define M($F$) as the number of models of the formula $F$, $m$ as the number of clauses in $F$, $n$ as the number of variables $F$ contains, Var($C$) as the variables in $C$, lit($C$) as the set of literals in the clause $C$, and lit ($C'$)-lit ($C$) as the set of literals that appear in the clause $C'$ but not in the clause $C$. We also use $F(\mu/\eta)$ to denote the substitution of $\mu$ by $\eta$ in the formula $F$, where $\mu$ is either a literal or a clause and $\eta$ is either a literal or *false*. To avoid a tedious enumeration of trivialities, if more than one literal is substituted by *false*, $\mu$ is usually expressed as a set of literals.

A formula $F$ in CNF can be expressed as an undirected graph called constraint graph. In the constraint graph $G$, the vertexes are the variables of $F$ and the edges between two vertexes if the corresponding variables appear together in some clause of $F$. A component of such a graph is a maximal sub-graph such that for every pair of vertices in the sub-graph, there is a path between the pair of vertices. Let $F_1, F_2, …, F_k$ be the sub-formulae of $F$ corresponding to the components. Then,

$$M(F)=M(F_1)\times M(F_2)\times … \times M(F_k) \qquad (1)$$

Given a formula $F$, the basic strategy of Davis-Putnam-Logemann-Loveland (DPLL) is to arbitrarily choose one or more variables that appear in $F$. That means that we branch on one or more variables in the formula $F$, i.e., we assign values to the variable(s) such that the problem for $F$ is reduced to the problem for two or more formulae.

**Estimating the Running Time**

In this section, we explain how to compute an upper bound on the running time of a DPLL-style algorithm. At first, we present a notion called branching tree. The branching tree (Hirsch 2000) is a hierarchical tree structure with a set of nodes, each of which is labeled with a formula. Suppose there is a node labeled with a formula $F$, then its sons labeled with $F_1$, $F_2, … , F_k$ are obtained by branching on one or more variables in the formula $F$, i.e., assigning values to the variable(s) such that the formula $F$ is reduced to two or more sub-formulae $F_1, F_2, … , F_k$ with fewer variables. Indeed, the construction of a branching tree can be viewed as an execution of a DPLL-style algorithm. Therefore, we use the branching tree to estimate the running time of our algorithm.

Suppose there is a branching tree whose nodes are labeled with formulae. To each node we attach a branching vector. Let us consider a node labeled with $F$ and its sons labeled with $F_1, F_2, …, F_k$. The branching vector of the node labeled with $F$ is a $k$-tuple $(r_1, r_2,…, r_k)$, where $r_i=f(F)-f(F_i)$ and $r_i >0$ ($1\leq i \leq k$ and $f(F)$ is the number of variables of $F$). From the definition of the branching vector of the node, it is easy to see that the branching tree requires that each sub-formula has smaller complexity than the initial formula after the formula being spitted. The characteristic polynomial of the branching vector is defined as follows:

$$h(x) = 1 - \sum_{i=1}^{k} x^{-r_i} \qquad (1)$$

The positive root of this polynomial is called the branching number, denoted by $\lambda (r_1, r_2,…, r_k)$. And we assume that the branching number of the leaves is 1. The maximum branching numbers of nodes is defined as the branching number of the branching tree, expressed by $\max \lambda (r_1, r_2,…, r_k)$. Actually, the branching number of a branching tree has an important relationship with the running time (T($n$)) of a DPLL-style algorithm. At first, assume the running time of a DPLL-style algorithm performing on each node is in polynomial time. Then the following inequality is obtained.

$$T(n) \leq (\max \lambda (r_1, r_2,…, r_k))^n \times \text{poly}(F) \qquad (2)$$

$$= (\max \sum_{i=1}^{k} T(n-r_i))^n \times \text{poly}(F)$$

where $n$ is the number of variables in $F$, ploy($F$) is the polynomial time executing on the node $F$, and

$$\lambda (r_1, r_2,…, r_k) = \sum_{i=1}^{k} T(n-r_i) \qquad (3)$$

In addition, if a #XSAT problem recursively solved by a DPLL-style algorithm, the time required doesn't increase, for

$$\sum_{i=1}^{k} T(n_i) \leq T(n) \text{ where } n = \sum_{i=1}^{k} n_i \qquad (4)$$

where $n$ is the number of variables, $n_i$ is the number of variables in the sub-formula $F_i$ ($1 \leq i \leq k$) of the formula $F$. Note that when analyzing the running time, we ignore the polynomial factor so that we assume that all polynomial time computations take $O(1)$ time in this paper.

## Algorithm for Solving #XSAT

In this section, we present the algorithm #XSAT for counting models of the exact satisfiability problem and prove an upper bound $O(1.1995^n)$. Firstly we address some principles used in this part.

## Reduction Principles

In this subsection, we concentrate on introducing the common literals principle and describing how to use the resolution principles in solving the #XSAT problem.

---

Function $unit(F[l])$
1. If there exists a clause $l \vee l_1 \vee l_2 \vee ... \vee l_n$ in $F$, then remove the clause $l \vee l_1 \vee l_2 \vee ... \vee l_n$ and the literals $l_1, l_2, ..., l_n$ from $F$.
2. If there exists a clause $\neg l \vee l'_1 \vee l'_2 \vee ... \vee l'_k$ in $F$, remove $\neg l$ from $\neg l \vee l'_1 \vee l'_2 \vee ... \vee l'_k$.
3. For $1 \leq i \leq n$ do $unit(F[\neg l_i])$.
4. Return $F$.

Figure 1: Function $unit(F[l])$

---

Before presenting the common variables principle, function $unit(F[l])$ in Figure 1 is addressed, which recursively executes the propagation. The function takes a formula $F$ and a literal $l$ being assigned *true* as input. The detailed process of the function is presented as follows. (1) Remove all clauses containing literal $l$ from $F$; (2) delete all literals occurring with $l$ from the other clauses; (3) delete all occurrences of the negation of literal $l$ from $F$; (4) perform the process as far as possible. Finally, the function returns a simplified formula. For ease in writing, the function $unit(F[l])$ is called $F[l]$ for short in the following sections.

Now we begin to present the common variables principle. Supposing an exact SAT formula $F$ contains at least two clauses having at least variables in common. Then we can calculate the models of $F$ as follows:

**Common Literals Principle**. Let $F=(C \vee C_1) \wedge (C \vee C_2) \wedge ... \wedge (C \vee C_k) \wedge F'$, where $C$, $C_1$, $C_2$, ..., $C_k$ are sub-clauses, and $Var(C) \cap Var(F')=\emptyset$. Then,

$$M(F)=|C| \times M(F''[x]) + M(F''[\neg x]) \text{ where } F''= \quad (5)$$

$(x \vee C_1) \wedge (x \vee C_2) \wedge ... \wedge (x \vee C_k) \wedge F'$ and $x \in C$

This principle is referred to as removing superfluous common literals. For example, $F=(x \vee y \vee z) \wedge (x \vee y \vee r) \wedge (s \vee \neg p)$. Then by executing the common literals principle, the formula can be reduced as $F''= (x \vee z) \wedge (x \vee r) \wedge (s \vee \neg p)$. It seems reasonable that $M(F)=M(F'')$ but this is not the case. In a model when $x=true$, $y$ is *false*. However, we know that there is a model that $x=false$ and $y=true$. This means that the reserved literal $x$ represents both, and therefore, we should hold the number of the common literals when counting the models of the #XSAT problem. Thus, from the above we know if a XSAT formula $F$ can be simplified by the common variables principle, then any two clauses in $F$ don't have more than two literals in common. And once there are two clauses having more than two literals in common, this means that $Var(C) \cap Var(F') \neq \emptyset$ and at least one common variable appears in another clause. Then we can branch on maximal degree of the common variables, which allows us to eliminate as many variables as possible.

Next we describe how to solve the #XSAT problem using the resolution principles which inspired by (Byskov et al. 2005). Actually, the resolution principle is a well-known technique for removing variables occurring both positive and negative and efficiently solving SAT problem. However, it is a pity that the technique can't be applied in solving #SAT for sometimes it eliminates a few variables which leads to the wrong result. Although #XSAT is also a counting problem, but the technique is suitable to solve #XSAT on account of the nature of the XSAT problem. Since the complexity of each node in the branching tree is strictly greatly than its son, we just use the resolution principle for $(1, 1^+)$-literals and $(2, 2)$-literals so that the resolution never increase the complexity. Now suppose that an exact SAT formula $F$ contains a $(1, 1^+)$-literal or a $(2, 2)$-literal, then we obtain the following principles.

**Resolution Principle 1.** Let $F = (x \vee C) \wedge (\neg x \vee C_1) \wedge (\neg x \vee C_2) \wedge \ldots \wedge (\neg x \vee C_k) \wedge F'$, where $x$ is a literal, $C_1$, $C_2$, and $C_k$ are sub-clauses in $F$. Then,

$$M(F) = M(F'') \qquad (6)$$

where $F'' = (C \vee C_1) \wedge (C \vee C_2) \wedge \ldots \wedge (C \vee C_k) \wedge F'$

**Resolution Principle 2.** Let $F = (x \vee C_1) \wedge (x \vee C_2) \wedge (\neg x \vee C_3) \wedge (\neg x \vee C_4) \wedge F'$, where $x$ is a literal, $C_1$, $C_2$, $C_3$, and $C_4$ are sub-clauses in $F$. Then,

$$M(F) = M(F'') \text{ where } F'' = (C_1 \vee C_3) \wedge \qquad (7)$$
$$(C_1 \vee C_4) \wedge (C_2 \vee C_3) \wedge (C_2 \vee C_4) \wedge F'$$

By performing the resolution principles, if the reduced formula $F$ contains a variable occurring both unnegated and negated, the literal consisting of the variable must be a $(2^+, 3^+)$-literal. According to the two resolution principles, we can obtain the follows.

**Theorem 1.** Given a formula $F$, if $F$ only contains $(1, 0)$-literals, $(2, 0)$-literals, $(3, 0)$-literals, and it has a clause $x \vee y$, then $x$ and $y$ can be both removed.

Proof. In order to make the formula *F true*, the clause $x \vee y$ must be set *true*. And since in each clause only one literal can be *true*, $\neg x = y$. Thus, $y$ is substituted by $\neg x$ and this makes $x$ a (1, 1)-literals, (2, 1)-literals, or (2, 2)-literals, which can be removed by the two resolution principles. Therefore, $x$ and $y$ can be both removed in this case. □

---

Function *Reduce*(F)
1. If there is a unit clause $(x)$, then $F=F[x]$.
2. If there is a clause of the form $(x \vee y)$ and val($y$) has a smaller degree, then $F=F(x/\neg y)$.
3. If there is a clause of the form $(x \vee x \vee C_1)$, then $F=F(x/false)$.
4. If there is a clause of the form $(x \vee \neg x \vee C_1)$, then $F= F(C_1/false)$.
5. If there are two clauses of the forms $(x \vee C_1)$ and $(y \vee C_1)$, where val($y$) has a smaller degree, then $F=F(x/y)$.
6. If there are two clauses $C$ and $C'$ such that $C \subset C'$, then $F= F((\text{lit}(C')-\text{lit}(C))/false)$.
7. If there are two clauses of the forms $(x \vee C_1)$ and $(C_1 \vee C_2)$, then the two clauses can be substituted by $(x \vee C_1)$ and $(\neg x \vee C_2)$.
8. If there is a clause $(C_1 \vee C_2)$, where $C_1$ only contains singletons, $C_2$ is composed by the non-singletons, and $x \in C_1$, then $F= F(C_1/x)$ and add $C_1$ to $R$.
9. If there are clauses $(x \vee C)$, $(\neg x \vee C_1)$, $(\neg x \vee C_2)$,…, $(\neg x \vee C_k)$, and other clauses in $F$ doesn't contain var($x$), then these clauses can be substituted by $(C \vee C_1)$, $(C \vee C_2)$,…, $(C \vee C_k)$.
10. If there are four clauses $(x \vee C_1)$, $(x \vee C_2)$, $(\neg x \vee C_3)$, $(\neg x \vee C_4)$, and other clauses in $F$ doesn't contain var($x$), then the four clauses can be substituted by $(C_1 \vee C_3)$, $(C_1 \vee C_4)$, $(C_2 \vee C_3)$, and $(C_2 \vee C_4)$.
11. If there are clauses $(C \vee C_1)$, $(C \vee C_2)$,…, $(C \vee C_k)$, and there is no common variables between the other clauses in $F$ and $C$, then these clauses can be substituted by $(x \vee C_1)$, $(x \vee C_2)$,…, $(x \vee C_k)$ and add $C$ to $R$, where $x \in C$.
12. Repeat the above steps until $F$ doesn't satisfy the above conditions.
13. Return $F$ and $R$.

Figure 2: Function *Reduce*

**Helpful Functions**

The subsection discusses some functions used for simplifying the formulae. The first function *Reduce*(F) in Figure 2 is to simplify the formula $F$ by recursively executing the common literal principle, the resolution principles, and some standard reductions used by (Kulikov 2005). It takes the formula $F$ as input and returns the reduced $F$ and a set of $R$ recording the eliminating sub-clauses. The reason why we use $R$ is that when using the common literal principle, the number of literals in the removed sub-clause *C* should be hold. In addition, given a XSAT formula *F*, if a clause contains more than one singleton, we can reserve one singleton and remove the superfluous ones. However, the eliminating sub-clause containing only singletons also causes the wrong result when counting the models of *F*. Therefore, we introduce the set of *R* to record the eliminating sub-clauses. According to the function, we obtain the Theorem 2.

**Theorem 2.** In a reduced XSAT formula, there are no unit clauses and 2-clauses; no clause has more than one singleton; if there is a variable occurring both unnegated and negated, the literal consisting of the variable must be a $(2^+, 3^+)$-literal; for all pairs of clauses, each has at least two variables that do not occur in the other; and if two clauses have more than two literals in common, at least one common literal appears in another clause.

Proof. Let us analyze the theorem one by one based on the function *Reduce*. The reduced formula containing no unit clauses and 2-clauses can be directly obtained by line 1 and 2 of the function. It is easy to see that the clauses in the reduced formula contain at most one singleton from line 8. If the reduced formula contains a variable occurring both unnegated and negated, then the literal composed of the variable may be a $(1, 1^+)$-literal, a $(2, 2)$-literal, or a $(2^+, 3^+)$-literal. When the literal is a $(1, 1^+)$-literal or a $(2, 2)$-literal, the literal can be removed by line 9 and 10. So in this case the literal must be a $(2^+, 3^+)$-literal. For all pairs of clauses, each having at least two variables that do not occur in the other can be obtained by line 7. In addition, if two clauses have more than two literals in common, then at least one common literal appearing in another clause can be obtained by line 11. □

Function $\Omega (F, \Phi)$
1. If $\Phi = \{l\}$ and $l$ is a literal in $C_1$ of $R$,
   then return $Reduce(F[l])$ and $S = |C_1|$.
2. If $\Phi = \{l\}$ and $l$ doesn't occur in $R$,
   then return $Reduce(F[l])$ and $S = 1$.
3. If $\Phi = \{l_1 \vee l_2\}$, then return $Reduce(F(l_1 / \neg l_2))$
4. If $\Phi = \{l_1, l_2\}$ and $l_1$ is a literal in $C_1$ of $R$,
   then return $Reduce((F[l_1])[l_2])$ and $S = |C_1|$.
5. If $\Phi = \{l_1, l_2\}$ and $l_1$ and $l_2$ don't occur in $R$,
   then return $Reduce((F[l_1])[l_2])$ and $S = 1$.

Figure 3: Function $\Omega$

The second function $\Omega (F, \Phi)$ in Figure 3 is to assign $\Phi$ *true* and reduce the formula after assigning values to the variables. The input to the function is the formula, and $\Phi$, where $\Phi$ can be a literal, two literals, or a clause consisting of only two literals. When $\Phi$ is a literal, the literal is set *true* directly; when $\Phi$ is a clause, the clause is fixed *true* and it implies that the only two literals have the complementary truth value; when $\Phi$ is two literals, the two literals are assigned *true* respectively.

Note that in this function, $S$ is an integer value recording the length of the eliminating sub-clause that $\Phi$ appears in. We assume that $P=1$ when the literals in $\Phi$ don't appear in $R$.

---

Algorithm #XSAT$_3$($F$)

Case 1: $F$ has an empty clause. Return 0.

Case 2: $F$ is empty. Return 0.

Case 3: $m \leq 4$. Return MC ($F$).

Case 4: $F$ contains disjoint components $F_1, F_2, \ldots, F_k$. Return #XSAT$_3$ ($F_1$) × #XSAT$_3$($F_2$) × … × #XSAT$_3$ ($F_k$).

Case 5: $F$ contains two clauses having at least two literals in common.
Pick two maximal degree common variables $x$ and $y$ and return #XSAT$_3$ ($\Omega$ ($F$, $\{x \vee y\}$)) + $S$ × #XSAT$_3$ ($\Omega$ ($F$, $\{\neg x, \neg y\}$)).

Case 6: If $F$ contains a 3-clause $x \vee y \vee z$, where $x$ isn't a singleton.
  1. If there is a 4-clause $x \vee p \vee q \vee r$ none of which is singleton, then return #XSAT$_3$ ( $\Omega$ ($F$, $\{x \vee p\}$)) + #XSAT$_3$ ($\Omega$ ($F$, $\{q \vee r\}$)).
  2. Otherwise, return $S$ × #XSAT$_3$ ($\Omega$ ($F$, $\{x\}$))+ $S$ × #XSAT$_3$ ($\Omega$ ($F$, $\{\neg x\}$)).

Case 7: If $F$ contains a 4-clause $x \vee y \vee z \vee p$ having no singletons, then return #XSAT$_3$ ( $\Omega$ ($F$, $\{x \vee y\}$)) + #XSAT$_3$ ($\Omega$ ($F$, $\{z \vee p\}$)).

Case 8: If $F$ contains a 4-clause $x \vee y \vee z \vee p$, where $p$ is a singleton and $z$ also occurs in other 4-clause, then return #XSAT$_3$ ( $\Omega$ ($F$, $\{x \vee y\}$)) + #XSAT$_3$ ( $\Omega$ ($F$, $\{z \vee p\}$)).

Case 9: If $F$ contains a 5-clause $x \vee y \vee z \vee p \vee q$ having no singletons, then return #XSAT$_3$ ($\Omega$ ($F$, $\{x \vee y\}$)) + #XSAT$_3$ ($\Omega$ ($F$, $\{z \vee p\}$)) + $S$ × #XSAT$_3$ ($\Omega$ ($F$, $\{q\}$)).

Case 10: If $F$ contains a 4-clause $x \vee y \vee z \vee p$, where $p$ is a singleton and $x$, $y$, and $z$ all occur in clauses of length at least 6, then return $S$ × #XSAT$_3$ ($\Omega$ ($F$, $\{x\}$)) + $S$ × #XSAT$_3$ ($\Omega$ ($F$, $\{y\}$) + #XSAT$_3$ ($\Omega$ ($F$, $\{z \vee p\}$)).

Case 11: If $F$ contains a clause of length at least 6, pick a literal $x$ that isn't a singleton in the clause and return $S$ × #XSAT$_3$($\Omega$ ($F$, $\{x\}$)) + $S$ × #XSAT$_3$ ($\Omega$ ($F$, $\{\neg x\}$)).

Case 12: If $F$ contains a 5-clause $x \vee y \vee z \vee p \vee q$, where $q$ is a singleton, then return #XSAT$_3$ ( $\Omega$ ($F$, $\{x \vee y\}$)) + $S$ × #XSAT$_3$ ($\Omega$ ($F$, $\{z\}$)) + #XSAT$_3$ ($\Omega$ ($F$, $\{p \vee q\}$)).

Figure 4: #XSAT$_3$ Algorithm

## Algorithm #XSAT

We propose a main algorithm #XSAT (Figure 5) which makes use of another algorithm $\#XSAT_3$ when $\varphi(F) \leq 3$. Both of the algorithms are based on the DPLL and take the reduced XSAT formula $F$ as the input. The basic idea of the two algorithms is to choose a variable and recursively count the number of satisfying assignments where the variable is *true* and the variable is *false*. At first, we present the framework of our algorithm $\#XSAT_3$ for solving #XSAT when $\varphi(F) \leq 3$ in Figure 4. Note that in the algorithm MC($F$) is a function that solves the #XSAT by exhaustive search. As we all know, if a #XSAT instance is solved by exhaustive search, it will spend a lot of time. However, when the number of clauses that the formula $F$ contains is so few, it may run in polynomial time. Therefore, we use the function MC($F$) only when the number of clauses isn't above 4, which can guarantee the exhaustive search runs in polynomial time. In addition, since the operation on each node is the function $\Omega(F, \Phi)$ running in polynomial time, we analyze the algorithms $\#XSAT_3$ and #XSAT using the measure described above in the following theorems.

**Theorem 3.** Algorithm $\#XSAT_3$ runs in $O(1.1995^n)$ time, where $n$ is the number of the variables.

Proof. Let us analyze the algorithm case by case.

Case 1, 2, and 3: These cases run in $O(1)$.

Case 4: This case doesn't increase the time needed.

Case 5: Suppose the two clauses are of the forms $(x \vee y \vee C)$ and $(x \vee y \vee C')$. We know that at least the literal $x$ appears in another clause based on Theorem 2. The running time $T(n)$ of the algorithm satisfies the recursive relation

$$T(n)=T(n-|C|-|C'|+|C \cap C'|-1)+T(n-2-i) \qquad (8)$$

where $i$ is an integer and $C \cap C'$ is the set of literals occurring in both sub-clauses. When $(x \vee y)=$*true*, the literals in $C$ and $C'$ are removed. When $x=$*false* and $y=$*false*, the two literals are also removed. Furthermore, once $|C|=2$ one literal is removed by function *Reduce* (line 2), i.e., $i=1$. And the same situation is encountered when $|C'|=2$. Here, since each clause has at least two variables that do not occur in the other by Theorem 2, $|C| \geq 2$, $|C'| \geq 2$, and $|C \cap C'| \leq$ min ($|C|-2, |C'|-2$). Moreover, if $|C \cap C'| \geq 1$, $x$ and $y$ both occur in another clauses based on Theorem 2. When $(x \vee y)=$*true*, $y$ is substituted by $\neg x$ and this makes $x$ a $(1, 1^+)$-literals, which we can remove by resolution principle 1. Therefore, it is easy to see that the worst case occurs when $|C|=3, |C'|=3, |C \cap C'|=0$ or $|C \cap C'|=1$. The time needed in this case is bounded by $T(n)=T(n-7)+T(n-2)$ with solution $O(1.1908^n)$.

Case 6.1: When $(x \vee p)=true$, the other two literals are fixed *false* so as to be removed. Besides, $p$ is substituted by $\neg x$ by function *Reduce* (line 2) and this makes $x$ a $(1, 1^+)$-literals, which we can remove by resolution principle 1. Therefore, the current formula contains at least four variables less than $F$. We have $T(n)=2T(n-4)$ because the same situation arises when $(q \vee r)=true$. This case takes $O(1.1892^n)$ time.

Case 6.2: Due to the previous cases, we know that the overlap of the clauses is only one literal and the literal is monotone. Suppose the other clause that $x$ occurs in is $C'$. The case $|C'|=3$ runs in $T(n)=T(n-5)+T(n-4)$ time and the remaining cases run in $T(n)=T(n-|C'|-2)+T(n-2-i)$, where $i$ is an integer. When $|C'|=4$ and $z$ is singleton, the length of the clause the literal $y$ in must be also 4 and there is a singleton in it. When $x=false$, it makes two singletons occur in the 4-clause and one of the singletons can be removed by function *Reduce* (line 8), i.e., $i=1$. Therefore, the worst running time is $T(n)=T(n-7)+T(n-2)$ with solution $O(1.1908^n)$.

Case 7: When $(x \vee y)=true$, the other two literals are fixed false so as to be removed. Besides, $y$ is substituted by $\neg x$ by function *Reduce* (line 2) and this makes $x$ a $(1, 1^+)$-literals, which we can remove by resolution principle 1. Therefore, the current formula contains at least four variables less than $F$. We have $T(n)=2T(n-4)$ because the same situation arises when $(z \vee p)=true$. This case takes $O(1.1892^n)$ time.

Case 8: This case is similar to case 7 when $(x \vee y)=true$ and it removes four variables. On account of the previous cases, we know that the other 4-clause that $z$ also occurs is a clause containing a singleton. Once $(z \vee p)=true$, $z$ is substituted by $\neg p$. This makes two singletons occur in the other 4-clause containing $z$ and one of the singletons can be removed by function *Reduce* (line 8). Therefore, the time needed in this case is bounded by $T(n)=2T(n-4)$ with solution $O(1.1892^n)$.

Case 9: The same circumstance as case 8 arises when $(x \vee y)=true$ and $(z \vee p)=true$. Five variables are removed due to $(x \vee y)=true$ and one more variable is removed when $x$ or $y$ occur in a 4-clause. When $p=true$, at least nine variables removed except the case when $p$ occurs in a 4-clause. Above all, it is easy to see that the worst case occurs when all the five literals in the 5-clauses also appear in other 5-clauses. Hence, the time needed in this case is bounded by $T(n)=T(n-9)+T(n-5)+T(n-5)$ with solution $O(1.1995^n)$.

Case 10: Since $x=true$, at least nine variables are removed. This is the same as $y=true$. Owing to $(z \vee p)=true$, three variables are removed. Therefore, the recurrence is $T(n)=T(n-9)+T(n-9)+T(n-3)$ with solution $O(1.1925^n)$.

Case 11: According to the above cases, the literal $x$ that isn't a singleton may occur in a 5-clause. So at least ten variables can be removed due to $x=true$. When $x=false$, at least the variable $x$ can be removed. The time needed is bounded by $T(n)=T(n-10)+T(n-1)$ with solution $O(1.1975^n)$.

Case 12: Each of the remaining clauses is 5-clause with one singleton based on the previous cases. The similar case as 7 when $(x \vee y)=true$ and this makes five variables remove. When $z=true$, at least nine variables are removed. In addition, five variables are removed because the similar case as 8 occurs when $(p \vee q)=true$. So the time needed in this case is bounded by $T(n)=T(n-9)+T(n-5)+T(n-5)$ with solution $O(1.1995^n)$.

In total, #XSAT$_3$ runs in $O(1.1995^n)$ time.  □

---

Algorithm #XSAT($F$)
Case 1: $F$ has an empty clause. Return 0.
Case 2: $F$ is empty. Return 0.
Case 3: $F$ contains disjoint components $F_1, F_2, \ldots, F_k$.
Return #XSAT $(F_1) \times$ #XSAT $(F_2) \times \ldots \times$ #XSAT $(F_k)$.
Case 4: $F$ contains a $(2^+, 3^+)$-literal $x$.
Return $S \times$ #XSAT $(\Omega (F, \{x\})) + S \times$ #XSAT $(\Omega (F, \{\neg x\}))$.
Case 5: $F$ contains two clauses having at least two literals in common.
Pick two maximal degree common variables $x$ and $y$ and return #XSAT$_3$ $(\Omega (F, \{x \vee y\})) + S \times$ #XSAT$_3$ $(\Omega (F, \{\neg x, \neg y \}))$.
Case 6: $\varphi (F) \geq 4$. Pick a variable $x$ such that $\varphi (x) \geq 4$.
Return $S \times$ #XSAT $(\Omega (F, \{x\})) + S \times$ #XSAT $(\Omega (F, \{\neg x\}))$.
Case 7: $\varphi (F) \leq 3$. Return #XSAT$_3(F)$.

Figure 5: #XSAT Algorithm

---

**Theorem 4.** Algorithm #XSAT runs in $O(1.1995^n)$ time, where $n$ is the number of the variables.

Proof. Let us analyze the algorithm case by case.

Case 1 and 2: These cases run in $O(1)$.

Case 3: This case doesn't increase the time needed.

Case 4: It is easy to see that the worst case occurs when the all variable $x$ occurs in the 3-clauses because in a reduced formula the length of the clause is at least three by Theorem 2. So when $x=true$, seven variables are removed; when $x=false$, five variables are removed. The time needed is bounded by $T(n)=T(n-7)+T(n-5)$ with solution $O(1.1238^n)$.

Case 5: This case is the same with the case 5 in Theorem 3. Therefore, the time needed in this case is bounded by $T(n)=T(n-7)+T(n-2)$ with solution $O(1.1908^n)$.

Case 6: Suppose $x$ occurs in $C_1$, $C_2$, $C_3$, and $C_4$. When $x=true$, the other literals in every clause containing $x$ are fixed *false* so that it makes $|C_1|+|C_2|+|C_3|+|C_4|-3$ variables remove. When $x=false$, one variable is removed and at least one more variables can be removed once $|C_i|=3$ ($i=1, 2, 3, 4$). Therefore, it is easy to see that the worst case occurs when $x$ occurs in the 4-clauses. So at least thirteen variables can be removed due to $x=true$. When $x=false$, at least the variable $x$ can be removed. The time needed is bounded by $T(n)=T(n-13)+T(n-1)$ with solution $O(1.1632^n)$.

Case 7: The result follows from Theorem 3 if $\varphi(F) \leq 3$.

In total, #XSAT runs in $O(1.1995^n)$ time. □

## Conclusion

This paper addresses the worst-case upper bound for #XSAT. The algorithm presented is DPLL-style algorithm. In order to improve the algorithms, we put forward a new common literals principle to simplify the formulae and firstly inject the resolution principles into solving #XSAT problem, which further improves the efficiency of the algorithm. After a skillful analysis of these algorithms, we obtain the worst-case upper bound $O(1.1995^n)$ time for #XSAT, where $n$ is the number of the variables.